\documentclass{article}

\usepackage[square,numbers]{natbib}
\usepackage[preprint]{nips_2018}

\usepackage[utf8]{inputenc}
\usepackage[T1]{fontenc}
\usepackage{hyperref}
\usepackage{url}
\usepackage{booktabs}
\usepackage{amsfonts}
\usepackage{nicefrac}
\usepackage{microtype}

\usepackage{amsmath}
\usepackage{mathtools}
\usepackage{algorithm}
\usepackage{algorithmicx}
\usepackage{algpseudocode}
\usepackage{multirow}
\usepackage{graphicx}
\usepackage{subcaption}
\usepackage{comment}

\title{Quantizing Convolutional Neural Networks for \\
  Low-Power High-Throughput Inference Engines
}
\author{
  Sean O.~Settle\thanks{} \\
  \texttt{sean.settle@xilinx.com} \\
  \And
  Manasa Bollavaram \\
  \texttt{manasab@xilinx.com} \\
  \And
  Paolo D'Alberto \\
  \texttt{paolod@xilinx.com} \\
  \And
  Elliott Delaye \\
  \texttt{elliott@xilinx.com} \\
  \And
  Oscar Fernandez \\
  \texttt{oscarfer@xilinx.com} \\
  \And
  Nicholas Fraser \\
  \texttt{nfraser@xilinx.com} \\
  \And
  Aaron Ng \\
  \texttt{aaronn@xilinx.com} \\
  \And
  Ashish Sirasao \\
  \texttt{asirasa@xilinx.com} \\
  \And
  Michael Wu \\
  \texttt{miwu@xilinx.com} \\
  \AND
  Xilinx, Inc. \\
  San Jose, CA 95124 \\
}

\begin{document}

\maketitle

\begin{abstract}
Deep learning as a means to inferencing has proliferated thanks to its
versatility and ability to approach or exceed human-level accuracy.  These
computational models have seemingly insatiable appetites for computational
resources not only while training, but also when deployed at scales ranging
from data centers all the way down to embedded devices.  As such, increasing
consideration is being made to maximize the computational efficiency given
limited hardware and energy resources and, as a result, inferencing with
reduced precision has emerged as a viable alternative to the IEEE 754 Standard
for Floating-Point Arithmetic.  We propose a quantization scheme that allows
inferencing to be carried out using arithmetic that is fundamentally more
efficient when compared to even half-precision floating-point.  Our
quantization procedure is significant in that we determine our quantization
scheme parameters by calibrating against its reference floating-point model
using a single inference batch rather than (re)training and achieve end-to-end
post quantization accuracies comparable to the reference model.
\end{abstract}

\section{Introduction}
Current state-of-the-art convolutional neural networks (CNNs) might trace
their roots back to LeNet-5 digit recognition \cite{lecun1998gradient}
and CIFAR-10 classification \cite{krizhevsky2012cuda}, but have
long since evolved into behemoths orders of magnitude wider and deeper as
exemplified by VGG~\cite{simonyan2014very}, with complex branching first
popularized by grouped convolution layers in
AlexNet~\cite{krizhenvsky2012imagenet}, inception layers in
GoogLeNet~\cite{szegedy2014going}, and residual layers in
ResNet~\cite{he2015deep}.  However, the competition to improve accuracies has
led to diminishing returns both in terms of training and inferencing.
Therefore, networks such as SqueezeNet~\cite{iandola2016squeezenet} and
MobileNet~\cite{howard2017mobilenets} were consciously crafted to strike a fine
balance between inference accuracies and compute and storage footprints.

Training and inferencing of CNNs were once markets exclusively owned by CPUs
thanks to their ubiquity and ease of programming, but then GPUs exploded into
the training market with the advent of NVIDIA's CUDA, Khronos' OpenCL, and
AMD's HIP combined with their plethora of single-precision floating-point units.
As network sizes increased the significance of a network's target data type on
its compute and storage requirements started gaining increasing attention due to
the fundamental energy and space costs to perform arithmetic operations and
execute load/store operations between various levels of memory caches
 \cite{horowitz2014computing, dally2015high}.  For example, 32-bit integer
additions cost about 4x more energy to perform compared to 8-bit integers,
multiplications cost about 16x more energy, and load/stores cost about 4x or
more energy depending on changes in cache locality.

The expense of these operations launched a hardware arms race to accelerate
reduced-precision arithmetic as a means to scale out for data centers and down
for embedded devices beginning with fixed-point data types and then continuing
to floating-point data types, for which FPGAs with their reprogrammable logic
can be reconfigured with custom circuit designs optimized for target
reduced-precision arithmetic while fixed instruction set architectures such as
CPUs and GPUs would have to emulate reduced-precision arithmetic not yet
hardened.

The remainder of this paper is structured as follows:
Section~\ref{sec:quantization_scheme} presents our quantization scheme,
Section~\ref{sec:quantization_procedure} explains our quantization procedure,
Section~\ref{sec:experiments} details our experiments, and
Section~\ref{sec:discussion} concludes with a discussion on the significance of
our findings.

\section{Quantization Scheme}\label{sec:quantization_scheme}
An ideal quantization scheme for CNNs is one whose basic arithmetic operations,
i.e., multiplication and addition, are widely available in hardware for at least
some common quantization scheme parameters and has relatively little overhead
when executing convolution layers.  In this section
we introduce our dynamic floating-point data types and some advanced operations
specific to CNNs that we refer back to in later sections.

\subsection{Data Type}
The suitability of a quantization scheme is often first judged based on its
range and precision most commonly centered about zero as a proxy to an error tolerance for a target
CNN's real-valued numerical ranges, which may vary widely even between adjacent
layers.  To strike such balance between range and precision, our quantization
procedure supports any numerical format that can be expressed as
$\alpha \times \beta$ according to the following dynamic floating-point format:
\begin{equation}
  \label{eqn:dynamic_floating_point_scheme}
  \underbrace{\text{scale}}_{\alpha \text{ -- real value, e.g., IEEE 754 floating-point value}}
    \times \underbrace{2^\text{exponent}
    \times \text{signed integer}}_{\beta \text{ -- reduced-precision floating-point value}}.
\end{equation}
More specifically, $\beta$---defined by quantization parameters $n$ and $p$
that respectively represent its bitwidth and number of significand bits
thereof, and variable bits $b_{n - 1}, \dots, b_1, b_0$---is represented by
sign and magnitude as
\begin{equation}
  \label{eqn:sign_and_magnitude}
  \beta =
  \begin{dcases}
    (-1)^{b_{n - 1}} \times \sum_{i = 0}^{p - 1} \left(b_i \times 2^i\right)
      & \qquad \text{if } \hat{e} = 0 \\
    2^{\hat{e} - 1} \times (-1)^{b_{n - 1}} \times \left(2^{p} + \sum_{i = 0}^{p - 1} \left(b_i \times 2^i\right)\right)
      & \qquad \text{if } \hat{e} > 0,
  \end{dcases}
\end{equation}
or by two's complement as
\begin{equation}
  \label{eqn:twos_complement}
  \beta =
  \begin{dcases}
    -b_{n - 1} \times \sum_{i = p}^{p + 1} \left(2^i\right) + \sum_{i = 0}^{p - 1} \left(b_i \times 2^i\right)
      & \qquad \text{if } \hat{e} = 0 \\
    2^{\hat{e} - 1} \times \left(-b_{n - 1} \times \sum_{i = p}^{p + 1} \left(2^i\right) + 2^{p} + \sum_{i = 0}^{p - 1} \left(b_i \times 2^i\right)\right)
      & \qquad \text{if } \hat{e} > 0,
  \end{dcases}
\end{equation}
where for both representations
\begin{equation}
  \label{eqn:exponent}
  \hat{e} =
  \begin{dcases}
    0
      & \qquad \text{if } p = n - 1 \\
    \sum_{i = p}^{n - 2} \left(b_i \times 2^{i - p}\right)
      & \qquad \text{if } p < n - 1.
  \end{dcases}
\end{equation}

\begin{figure}[H]
\centering
\includegraphics[width=0.8\textwidth,keepaspectratio]{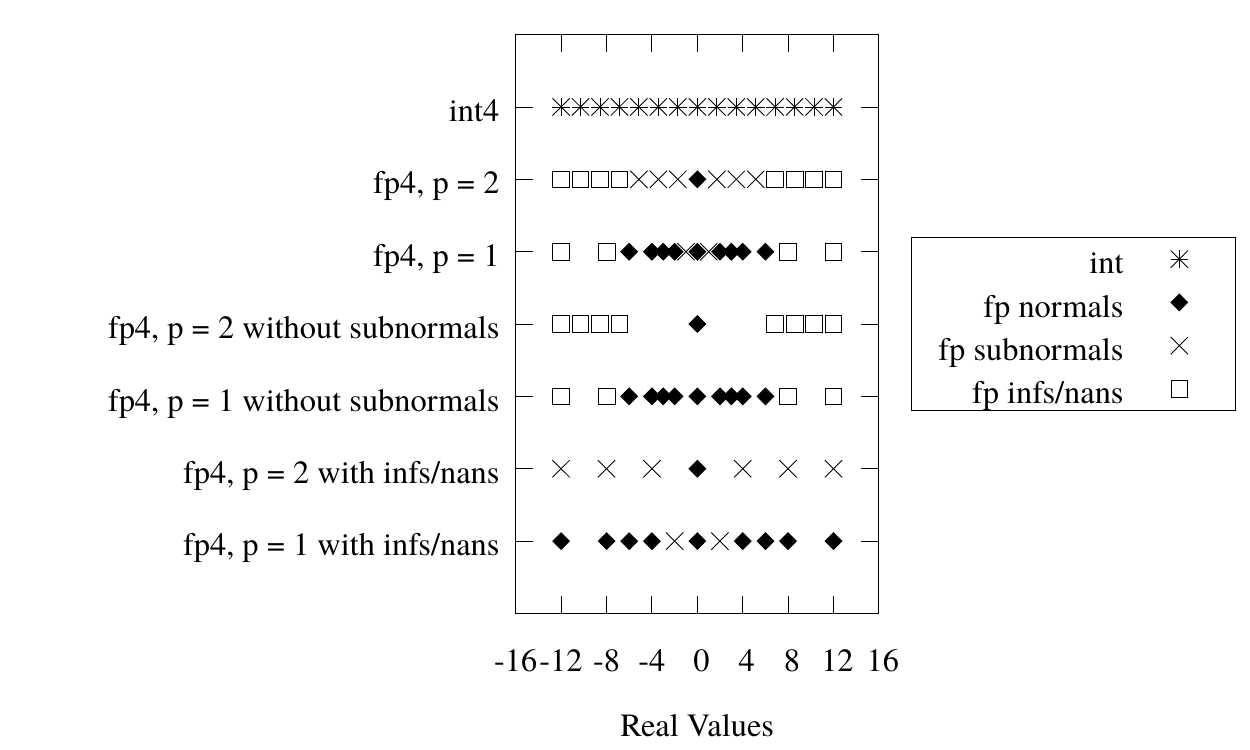}
\caption{Example quantization schemes normalized to the same range}
\label{fig:range_and_precision_normalized_to_the_same_range}
\end{figure}

With Equations~\ref{eqn:dynamic_floating_point_scheme},
\ref{eqn:sign_and_magnitude}, and \ref{eqn:exponent}
now being defined, the relationship between range and precision may be more
readily understood from the examples depicted in
Figure~\ref{fig:range_and_precision_normalized_to_the_same_range}.  In fact,
our dynamic floating-point quantization scheme is the floating-point analogue
to the dynamic fixed-point quantization scheme \cite{courbariaux2014training},
and a generalization and extension of the IEEE 754 Standard for Floating-Point
Arithmetic \cite{ieee2008ieee}, including default support for subnormal
values, i.e., $\hat{e} = 0$ and $\beta \ne 0$, in addition to normalized
values, but with the default option to extend numeric values in place of
infinity and NaN (quiet or signaling).

As highlighted by the normalization in
Figure~\ref{fig:range_and_precision_normalized_to_the_same_range}, it is often
more natural to frame the quantization scheme in terms of a threshold $\gamma$
using the simple conversion $\gamma = \alpha \times \beta_{max}$, or more
explicitly
\begin{equation}
  \label{eqn:threshold}
  \beta_{max} \leq
  \begin{dcases}
    2^{n - 1} - 1
      & \qquad \text{if } p = n - 1 \\
    2^{2^{n - p - 1} - 2} \times \left(2^{p + 1} - 1\right)
      & \qquad \text{if } p < n - 1,
  \end{dcases}
\end{equation}
where equality comes from using the full range by excluding Infs/NaNs.  In
other words, scale $\alpha$ represents the smallest positive subnormal value if
subnormals were included, and threshold $\gamma$ represents the largest
positive value allowed in the quantization scheme.

\subsection{Operations}
Basic arithmetic operations, that is binary operations between dynamic
floating-point data types, are just simple cases of more advanced operations
found in CNNs.  In this section we define two advanced operations that in turn
also define binary multiplication and addition.

If given dynamic floating-point numbers $x_i = \alpha_x \times \beta_{x_i}$
and $y_i = \alpha_y \times \beta_{y_i}$ such that $n_x = n_{x_i}$,
$n_y = n_{y_i}$, $p_x = p_{x_i}$, and $p_y = p_{y_i}$ for all $i$, then the
multiplication and accumulation found in convolution and inner product layers
is expressed as follows:
\begin{equation}
  \label{eqn:multiplication_and_accumulation}
  z = \alpha_z \times \beta_z
    = \sum_{i = 0}^{N - 1} \left(x_i \times y_i\right)
    = \alpha_z \times \frac{\alpha_x \times \alpha_y}{\alpha_z} \times \sum_{i = 0}^{N - 1} \left(\beta_{x_i} \times \beta_{y_i}\right).
\end{equation}
Now suppose that $\left[a_x, b_x\right]$ and $\left[a_y, b_y\right]$ are ranges
with the smallest upper-bounds such that they contain every $\beta_{x_i}$ and
$\beta_{y_i}$, respectively, where $b_x \ge 1$, $b_y \ge 1$,
$a_x = -b_x - \epsilon_x$, $a_y = -b_y - \epsilon_y$, and
$\epsilon_x, \epsilon_y \in \mathbb{Z}_{\ge 0}$.  Without loss of generality
let $a_y \times b_x \le a_x \times b_y$, and since by construction
$a_x \times a_y \ge b_x \times b_y$, then 
\begin{equation}
  \label{eqn:multiplication_and_accumulation_range}
  \sum_{i = 0}^{N - 1} \left(\beta_{x_i} \times \beta_{y_i}\right)
    \in \left[N \times a_y \times b_x, N \times a_x \times a_y\right].
\end{equation}
We can compute the sum of products above using a $q$-bit two's complement
signed-integer accumulator without loss of accuracy so long as
$-2^{q - 1} \le N \times a_y \times b_x$ and
$N \times a_x \times a_y \le 2^{q - 1} - 1$.
Therefore the smallest $q$ is
\begin{equation}
  \label{eqn:multiplication_and_accumulation_q}
  q =
    \left\lceil\log_2\left(N \times \left(b_x + \epsilon_x\right) \times \left(b_y + \epsilon_x\right) + 1\right) + 1\right\rceil
\end{equation}
and the largest $N$ is
\begin{equation}
  \label{eqn:multiplication_and_accumulation_N}
  N =
    \left\lfloor
      \left(2^{q - 1} - 1\right) / \left(\left(b_x + \epsilon_x\right) \times \left(b_y + \epsilon_y\right)\right)
    \right\rfloor,
\end{equation}
which generalizes a previous fixed-point analysis \cite{fu8-bit}, and indicates
that we should require $\epsilon_x = 0$ and $\epsilon_y = 0$ by always clamping
IEEE floating-point values to $\left[-\gamma, \gamma\right]$ and
reduced-precision floating-point values to
$\left[-\beta_{max}, \beta_{max}\right]$.

Next, if given dynamic floating-point numbers
$x_i = \alpha_{x_i} \times \beta_{x_i}$ such that $n_x = n_{x_i}$ and
$p_x = p_{x_i}$ for all $i$, then the addition found in eltwise addition layers
is expressed as follows:
\begin{equation}
  \label{eqn:addition}
  z = \alpha_z \times \beta_z
    = \sum_{i = 0}^{N - 1} x_i
    = \alpha_z \times \sum_{i = 0}^{N - 1} \left(\frac{\alpha_{x_i}}{\alpha_z} \times \beta_{x_i}\right).
\end{equation}
Now suppose that $\left[a_x, b_x\right]$ is a range with the smallest
upper-bound such that it contains every $\beta_{x_i}$, where $b_x \ge 1$,
$a_x = -b_x - \epsilon_x$, $\epsilon_x \in \mathbb{Z}_{\ge 0}$, and
$\alpha_{x_i} \le \alpha_z$, then
\begin{equation}
  \label{eqn:addition_range}
  \sum_{i = 0}^{N - 1} x_i \in \left[N \times a_x, N \times b_x\right].
\end{equation}
We can compute the sum above using a $q$-bit two's complement signed-integer
accumulator without loss of accuracy so long as
$-2^{q - 1} \le N \times a_x$ and $N \times b_x \le 2^{q - 1} - 1$.
Therefore the smallest $q$ is
\begin{equation}
  \label{eqn:addition_q}
  q =
  \begin{dcases}
    \left\lceil\log_2\left(N \times b_x + 1\right) + 1\right\rceil
      & \qquad \text{if } \epsilon_x = 0 \\
    \left\lceil\log_2\left(N \times b_x + N \times \epsilon_x\right) + 1\right\rceil
      & \qquad \text{if } \epsilon_x \ge 1
  \end{dcases}
\end{equation}
and the largest $N$ is
\begin{equation}
  \label{eqn:addition_N}
  N =
  \begin{dcases}
    \left\lfloor
      \left(2^{q - 1} - 1\right) / b_x
    \right\rfloor
      & \qquad \text{if } \epsilon_x = 0 \\
    \left\lfloor
      2^{q - 1} / \left(b_x + \epsilon_x\right)
    \right\rfloor
      & \qquad \text{if } \epsilon_x \ge 1,
  \end{dcases}
\end{equation}
which further supports our decision to require $\epsilon_x = 0$.

\section{Quantization Procedure}\label{sec:quantization_procedure}
An ideal quantization procedure for CNNs is one that with little effort can be
made backwards compatible with existing hardware and software.  Additionally,
it is preferential to be able to reuse existing models to save the time of
training from scratch, as featured in Ristretto~\cite{gysel2016hardware, gysel2018ristretto} and
Google's TensorFlow~\cite{jacob2017quantization}.  In fact, one would rather
not retrain as even that can be time consuming or require the entire labeled
data set which may not be available.  Another method found in Google's
TensorFlow determines some 8-bit fixed-point quantization parameters online for
each batch during inferencing \cite{tensorflow2017}, though it does have hooks
to save and load these parameters offline.  Elsewhere, NVIDIA's TensorRT
determines 8-bit fixed-point quantization parameters offline by inferencing
with an unlabeled calibration set and at each layer analyzing the difference
between the reference floating-point and fixed-point probability distribution
functions \cite{migacz20178-bit}.  Our method supports any dynamic
floating-point scheme and a unified offline/online quantization flow with
additional enhancements we found necessary for end-to-end post quantization
accuracies.  In order to limit the scope of this paper, we present our
quantization procedure as it pertains to offline quantization.

In our method an IEEE 754 floating-point value in a tensor is converted to its
reduced-precision floating-point value $\beta$ by first clamping the IEEE 754
floating-point value to a strictly symmetric threshold range
$[-\gamma, \gamma]$.  The threshold value $\gamma$ may be directly chosen from
some statistical analysis of the aforementioned tensor such as a percentile or
number of standard deviations, or from similar analyses of historically
comparable tensors as shown in Algorithm~\ref{alg:thresholding} using a
weighted average of over a dozen measures \cite{dalberto2009non-parametric},
though others have used just the Kullback-Leibler-I with reported success
\cite{migacz20178-bit}.  Next, the clamped value is divided by its scale
$\alpha$, then the result proceeds through several more steps in order to round
to the desired $\beta$.  A reduced-precision floating-point value $\beta$ is
converted to an IEEE 754 floating-point value by simply multiplying through by
its scale $\alpha$.

\begin{algorithm}[h]
\centering
\caption{Thresholding dynamic floating-point values}
\label{alg:thresholding}
\begin{algorithmic}[1]
\Function{threshold}{$x, n, p$}
  \State $pdf, bin\_edges \gets hist(abs(x)); cdf \gets cumsum(pdf)$
  \State $\delta \gets \infty; \gamma \gets bin\_edges[-1]$
  \For {$i \gets pow(2, n - 1) + 1, len(bin\_edges) - 1$}
    \State $r \gets cdf; r[i-1:] \gets 1$
    \State $r\_interp \gets interp(fpspace(0., 1., n, p), linspace(0., 1., i), r[:i])$
    \State $q\_interp \gets interp(linspace(0., 1., i), fpspace(0., 1., n, p), r\_interp)$
    \State $q \gets r$; $q[:i] \gets q\_interp$
    \State $\delta_{tmp} \gets measure(cdf, q); \gamma_{tmp} \gets bin\_edges[i])$
    \If {$\delta > \delta_{tmp}$}
      \State $\delta \gets \delta_{tmp}; \gamma \gets \gamma_{tmp}$
    \EndIf
  \EndFor
  \State \Return $\gamma, \delta$
\EndFunction
\end{algorithmic}
\end{algorithm}

We begin our quantization procedure with a few preprocessing steps to prepare
the network.  Where possible we merge adjacent fork/join memory operations,
e.g., multiple concat layers may be combined into a single concat layer.
Likewise, any adjacent layers that are linear operations, e.g., batchnorm,
scale, bias, convolution, and inner product layers, are combined into either a
convolution or inner product layer.  If an adjacent convolution or inner
product layer does not exist, an identity convolution or inner product layer is
created and inserted accordingly.  Splicing in identity operations such as the
aforementioned layers is a technique we heavily use to give us the ability to
explicitly downscale reduced-precision floating-point values during end-to-end
post quantization execution by factors we empirically find during the remainder
of our quantization procedure to be on the interval $(0, 1]$.  Where these
factors are determined to be unity, i.e., remain identity operations, then
those layers are spliced back out.  We apply this technique on edges directly
connecting fork layers to join layers, and before join layers that directly, or
indirectly, could not otherwise be explicitly downscaled, e.g., a maxpool layer
following an eltwise layer.

Once the preprocessing steps are complete we iterate through the network
quantizing weights with threshold values equal to the maximum absolute value
per respective output channel.  Meanwhile, threshold values for activations are
measured in a bootstrap quantization fashion instead of a divide and conquer
approach.  Note that the choice of thresholds for biases may either be a
function of input thresholds and weight thresholds or output thresholds
depending on the target hardware implementation.  Furthermore, the
determination of threshold values for activations is delayed as long as
possible so as to not prematurely drop least significant bits causing
irreversible loss of information.  For example, the threshold of a convolution
layer--as specified by its associated scaling factor in
Equation~\ref{eqn:multiplication_and_accumulation}--followed by a ReLU layer is
not determined until after the ReLU layer.  In particular, this lazy
thresholding is necessary to match the scaling factors associated with each
input to joining layers such as concat and eltwise layers as indicated in
Equation~\ref{eqn:addition}.

\section{Experiments}\label{sec:experiments}
We conducted three sets of dynamic floating-point quantization experiments
that each target a variety of modern network features, mainly inception layers,
residual layers, and grouped convolution layers in GoogLeNet v1~\cite{szegedy2014going, guadarrama2014bvlc}, ResNet-50~\cite{he2015deep, he2016deep}, and
MobileNet v1~\cite{howard2017mobilenets, yang2017mobilenet}, which we will henceforth simply refer to as GoogLeNet, ResNet, and MobileNet, respectively.  Each of these models were trained by their authors on the ImageNet data set using
single-precision floating-point.  We chose GoogLeNet and ResNet as they
represent networks typically deployed in data centers, and MobileNet because it
is more indicative of networks designed for embedded devices.  For each set of
experiments we performed our quantization procedure for a calibration set
chosen randomly without replacement from the training images and then measured
the Top-1 inference accuracies for each set of experiments against 1600
validation images.  Throughout the three sets of experiments we also varied
the calibration set size between between 8, 32, and 128 to measure its relative
effect on accuracies.  Finally, unless otherwise mentioned, we include subnormals
but exclude Infs/NaNs from our quantization scheme.

\subsection{Subnormals and Infs/NaNs}
\begin{figure}[H]
\centering
  \begin{subfigure}[b]{0.4\textwidth}
  \centering
  \includegraphics[scale=0.4]{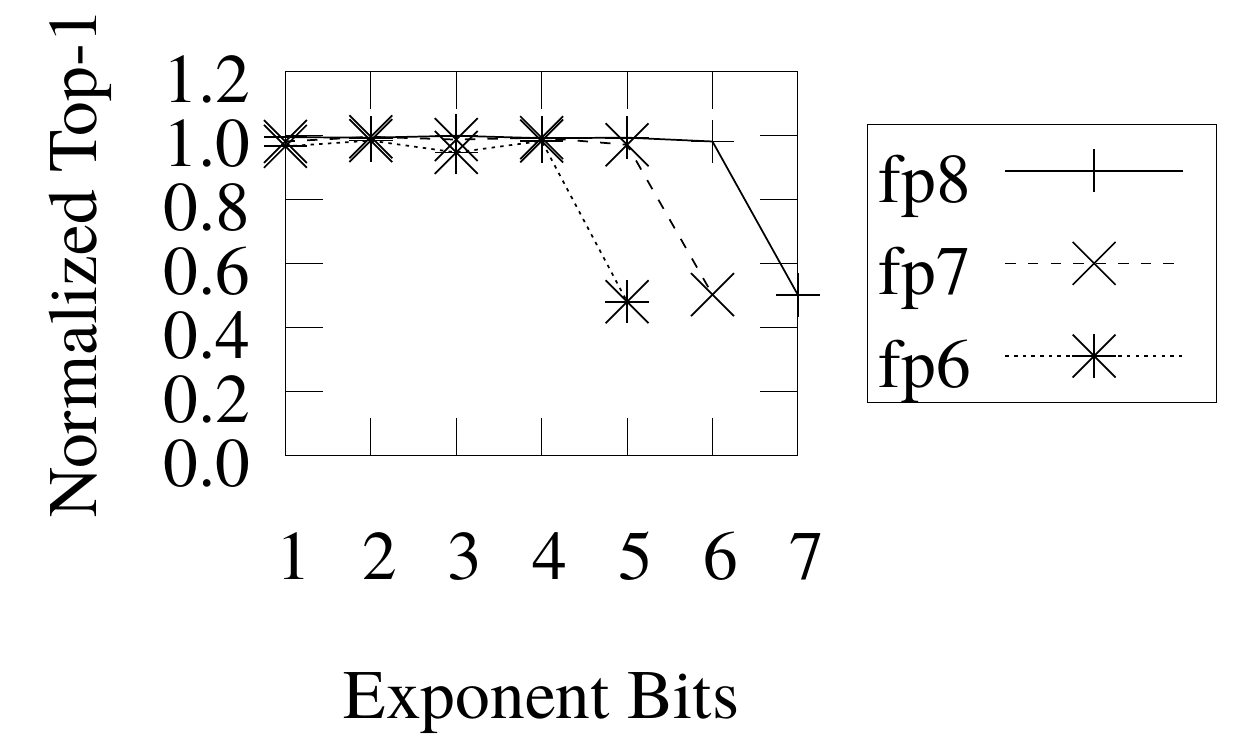}
  \caption{GoogLeNet}
  \label{fig:with_subnormals_but_without_infs_nans-googlenet}
  \end{subfigure}%

  \begin{subfigure}[b]{0.4\textwidth}
  \centering
  \includegraphics[scale=0.4]{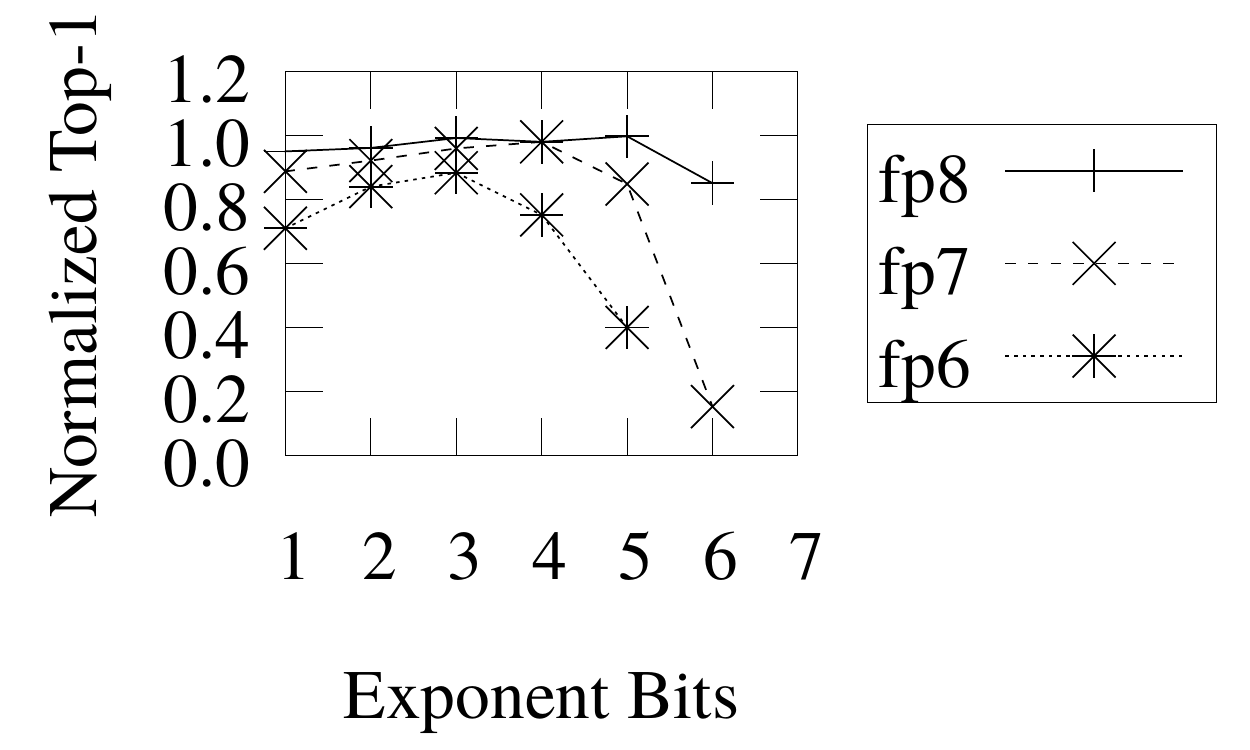}
  \caption{ResNet}
  \label{fig:with_subnormals_but_without_infs_nans-resnet}
  \end{subfigure}%
~
  \begin{subfigure}[b]{0.4\textwidth}
  \centering
  \includegraphics[scale=0.4]{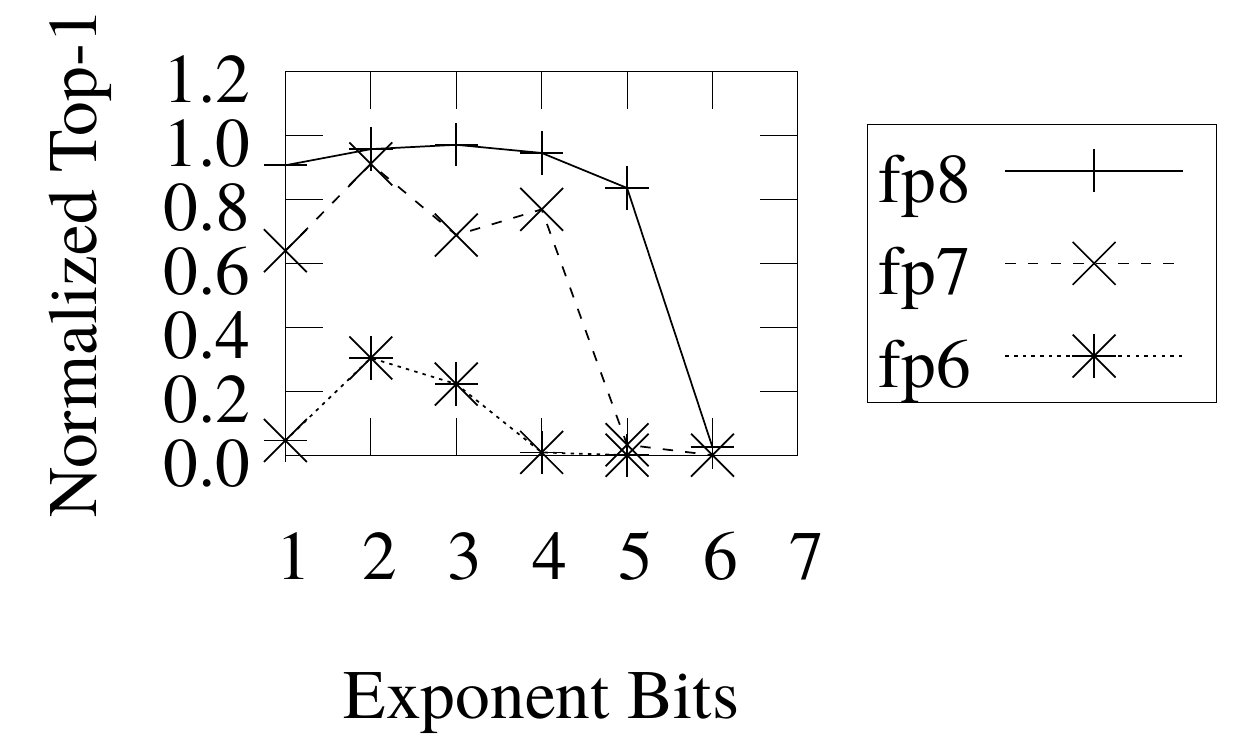}
  \caption{MobileNet}
  \label{fig:with_subnormals_but_without_infs_nans-mobilenet}
  \end{subfigure}
\caption{With subnormals but without Infs/NaNs, calibrated against 128 training images}
\label{fig:with_subnormals_but_without_infs_nans}
\end{figure}

\begin{figure}[H]
\centering
  \begin{subfigure}[b]{0.4\textwidth}
  \centering
  \includegraphics[scale=0.4]{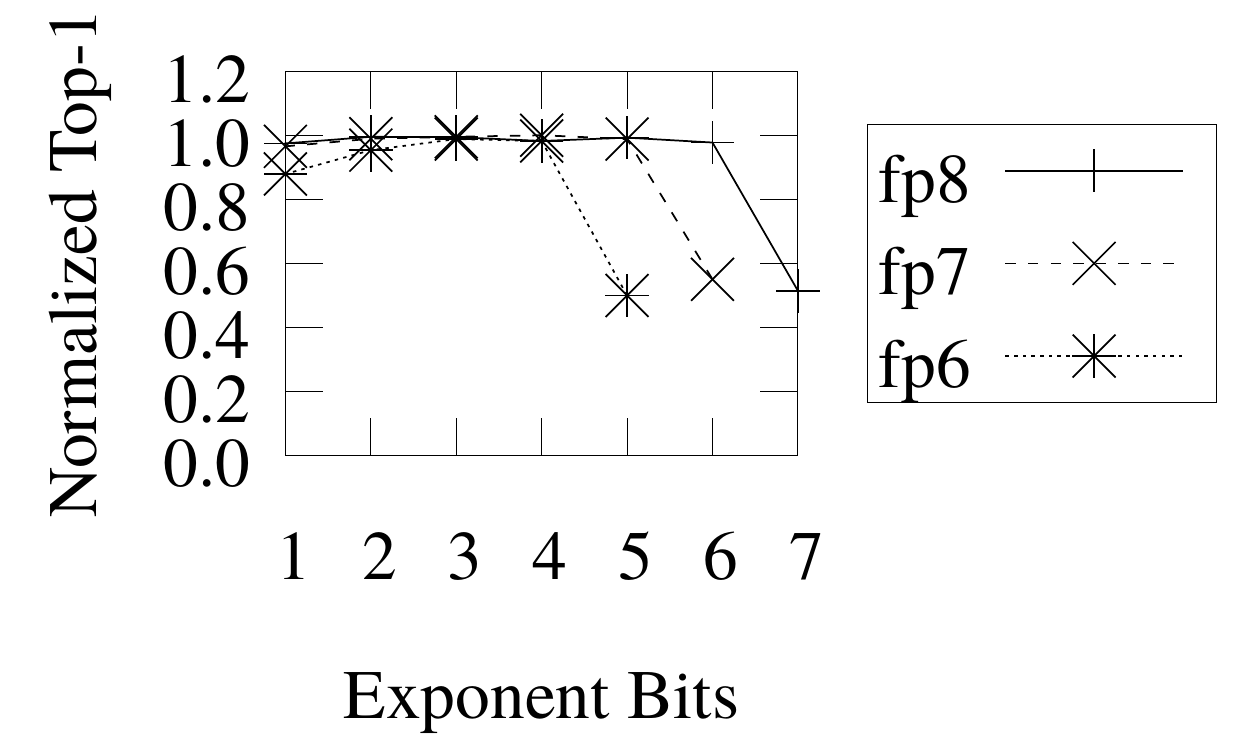}
  \caption{GoogLeNet}
  \label{fig:with_subnormals_and_infs_nans-googlenet}
  \end{subfigure}%

  \begin{subfigure}[b]{0.4\textwidth}
  \centering
  \includegraphics[scale=0.4]{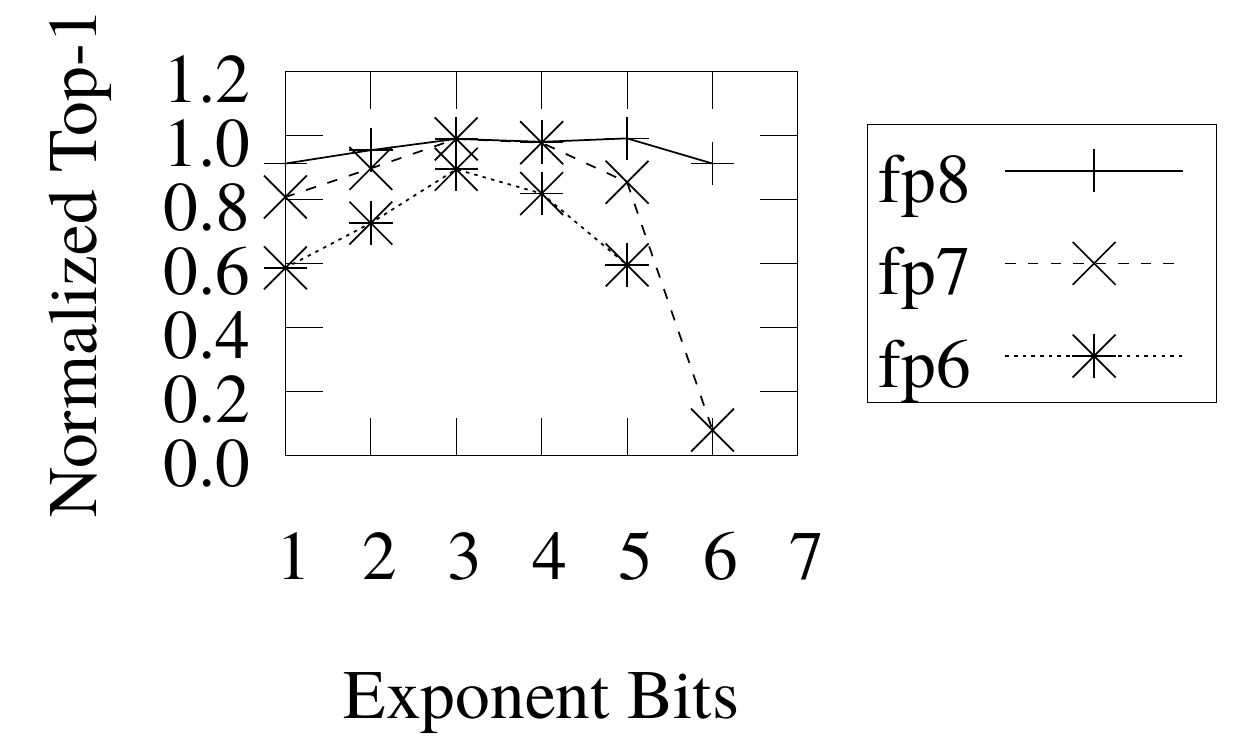}
  \caption{ResNet}
  \label{fig:with_subnormals_and_infs_nans-resnet}
  \end{subfigure}%
~
  \begin{subfigure}[b]{0.4\textwidth}
  \centering
  \includegraphics[scale=0.4]{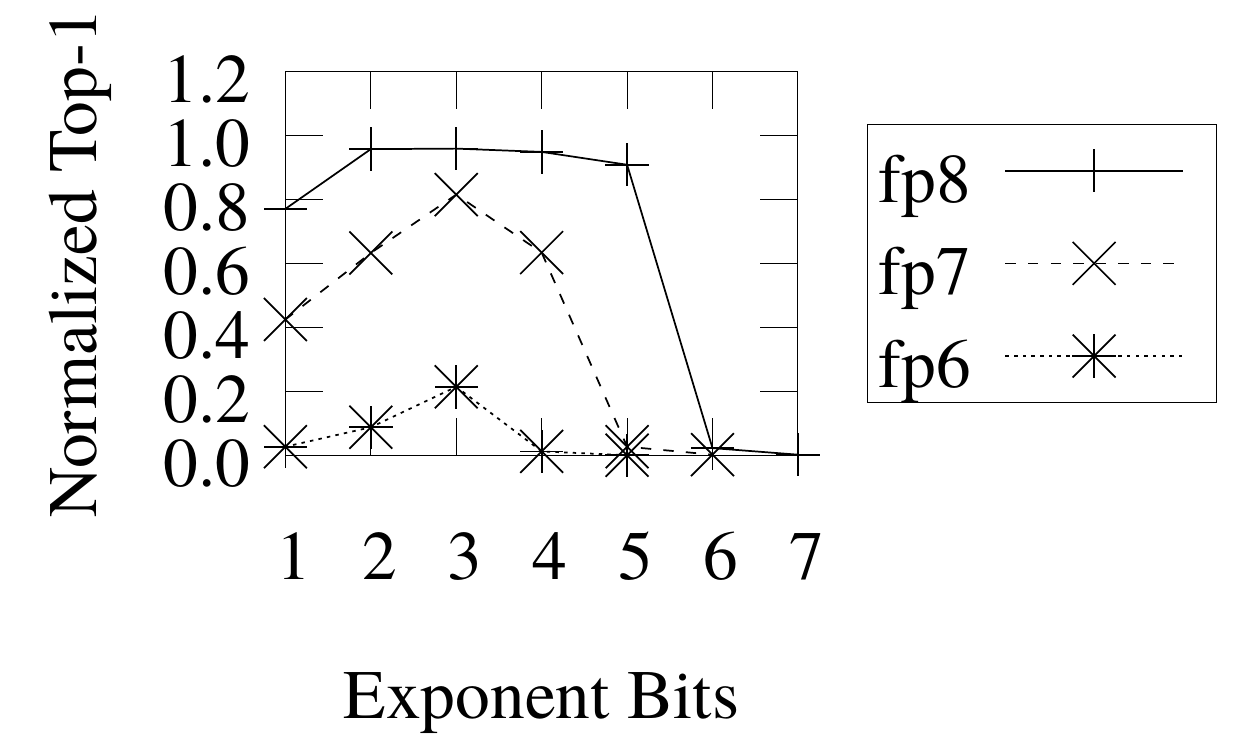}
  \caption{MobileNet}
  \label{fig:with_subnormals_and_infs_nans-mobilenet}
  \end{subfigure}
\caption{With subnormals and Infs/NaNs, calibrated against 8 training images}
\label{fig:with_subnormals_and_infs_nans}
\end{figure}

\begin{figure}[H]
\centering
  \begin{subfigure}[b]{0.4\textwidth}
  \centering
  \includegraphics[scale=0.4]{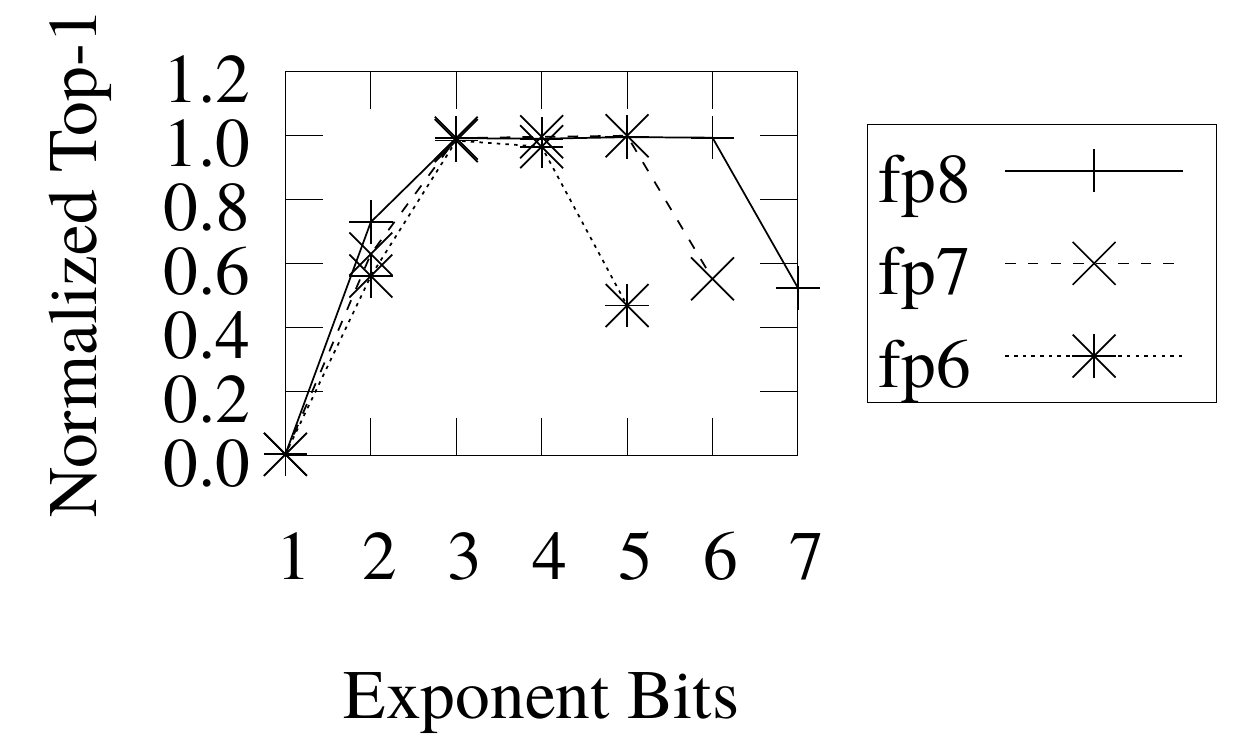}
  \caption{GoogLeNet}
  \label{fig:without_subnormals_or_infs_nans-googlenet}
  \end{subfigure}%

  \begin{subfigure}[b]{0.4\textwidth}
  \centering
  \includegraphics[scale=0.4]{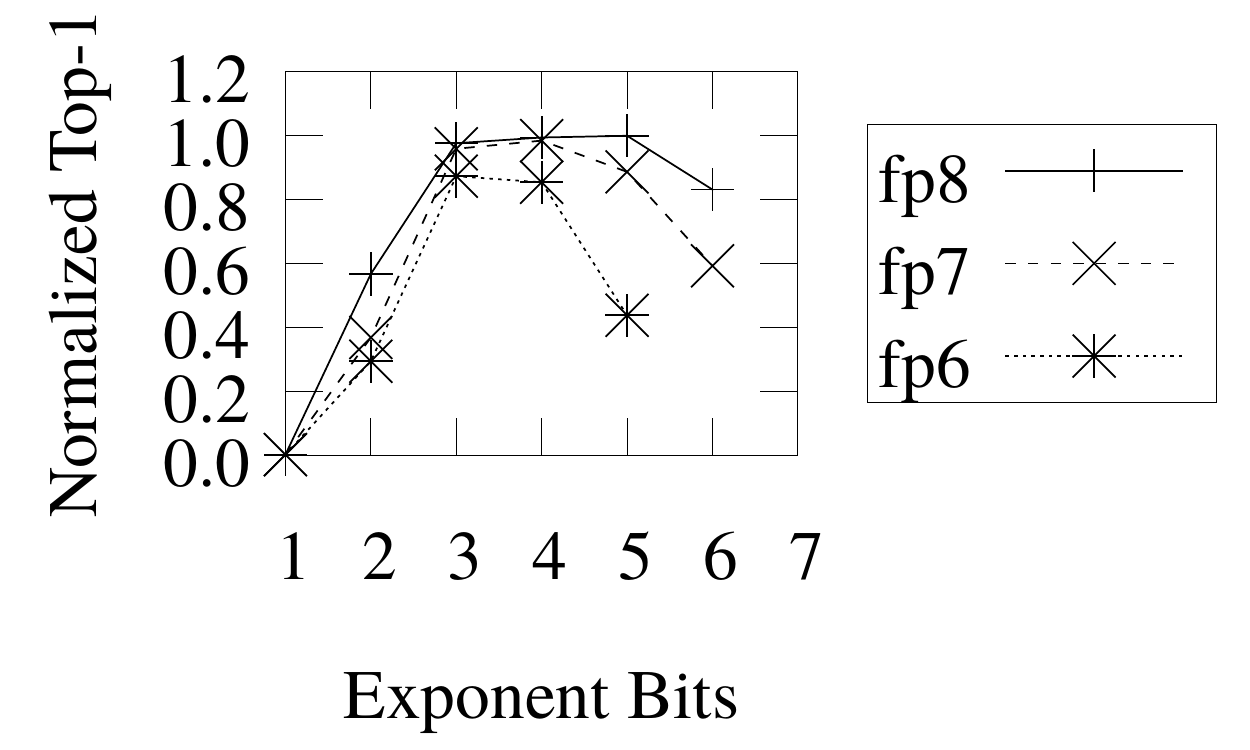}
  \caption{ResNet}
  \label{fig:without_subnormals_or_infs_nans-resnet}
  \end{subfigure}%
~
  \begin{subfigure}[b]{0.4\textwidth}
  \centering
  \includegraphics[scale=0.4]{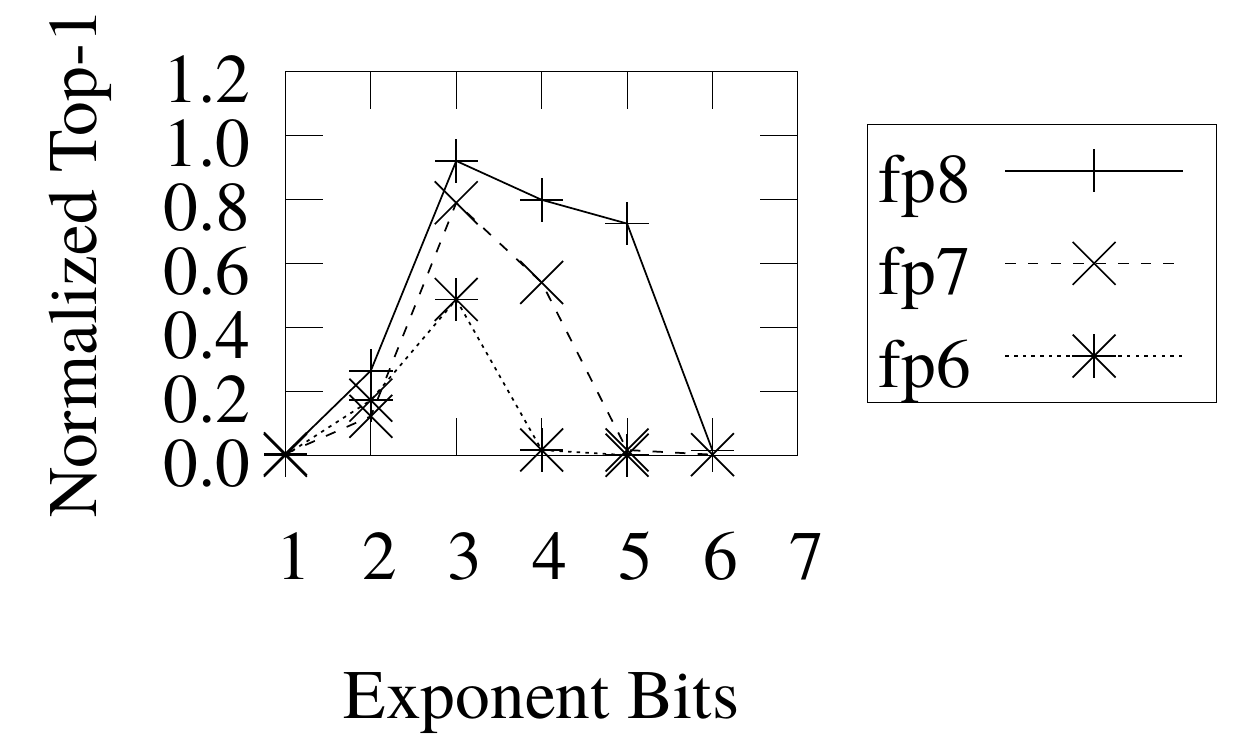}
  \caption{MobileNet}
  \label{fig:without_subnormals_or_infs_nans-mobilenet}
  \end{subfigure}
\caption{Without subnormals or Infs/NaNs, calibrated against 8 training images}
\label{fig:without_subnormals_or_infs_nans}
\end{figure}

The first set of experiments measure the effect of excluding subnormals or
including Infs/NaNs since both reduce the numeric values representable by our
quantization scheme, see
Figures~\ref{fig:with_subnormals_but_without_infs_nans},
\ref{fig:with_subnormals_and_infs_nans}, and
\ref{fig:without_subnormals_or_infs_nans}.  Here we use the same
quantization scheme for weights as we do for activations.  For the same
bitwidth, Figure~\ref{fig:with_subnormals_and_infs_nans} shows no discernible
difference in accuracies compared to
Figure~\ref{fig:with_subnormals_but_without_infs_nans}, while
Figure~\ref{fig:without_subnormals_or_infs_nans} exhibits a significant drop in
accuracies with fewer than 3 exponent bits.  In fact, in each experiment there
is no significant improvement in accuracies with more than 3 exponent bits.
Finally, compared to results for a single exponent bit in
Figure~\ref{fig:with_subnormals_but_without_infs_nans}, which are by definition
dynamic fixed-point schemes, there is between a 1--30\% improvement in
accuracies when using dynamic floating-point schemes of the same bitwidth and 3
exponent bits.  This improvement is more pronounced in MobileNet, less so in
ResNet, negligible in GoogLeNet, and generally more significant for smaller
bitwidths.

\subsection{Activations and Weights}
\begin{figure}[H]
\centering
  \begin{subfigure}[b]{0.4\textwidth}
  \centering
  \includegraphics[scale=0.4]{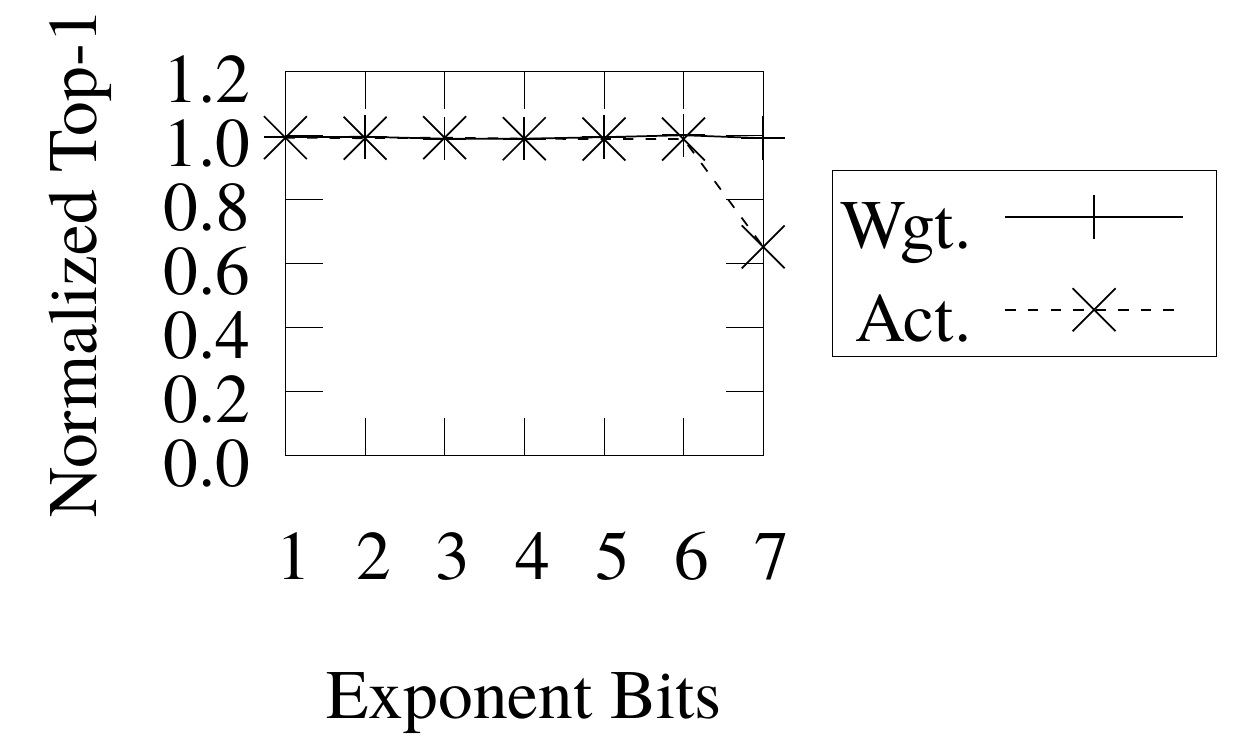}
  \caption{GoogLeNet}
  \label{fig:activations_and_weights-googlenet}
  \end{subfigure}%

  \begin{subfigure}[b]{0.4\textwidth}
  \centering
  \includegraphics[scale=0.4]{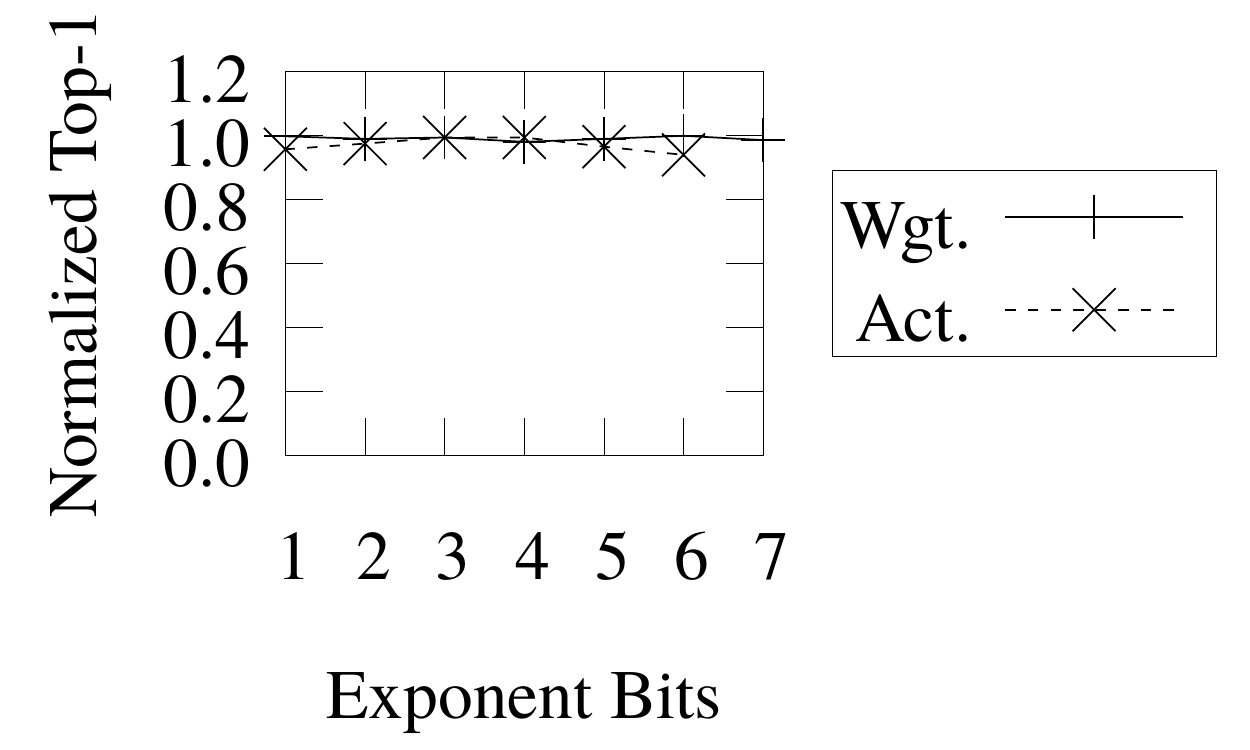}
  \caption{ResNet}
  \label{fig:activations_and_weights-resnet}
  \end{subfigure}%
~
  \begin{subfigure}[b]{0.4\textwidth}
  \centering
  \includegraphics[scale=0.4]{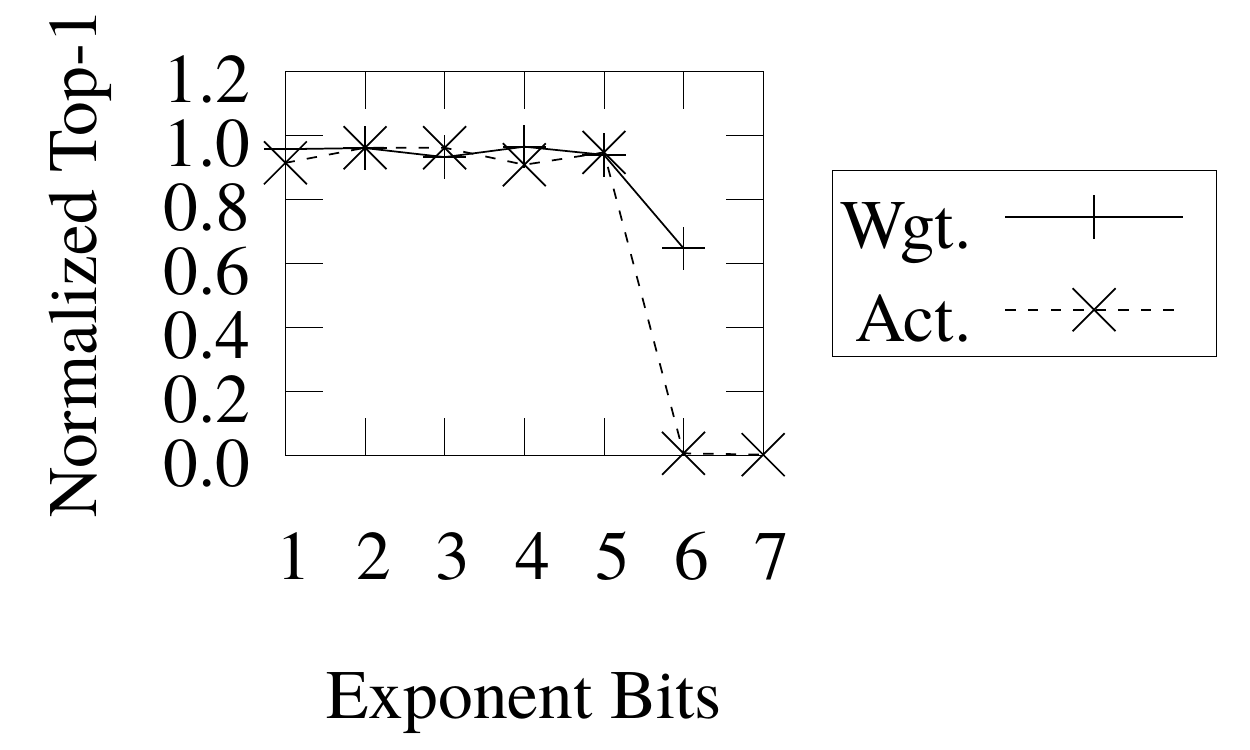}
  \caption{MobileNet}
  \label{fig:activations_and_weights-mobilenet}
  \end{subfigure}
\caption{Activations and weights based on $n = 8$, $p = 4$, calibrated against 32 training images}
\label{fig:activations_and_weights}
\end{figure}

The second set of experiments measure the effect of trading exponent bits for
significand bits on the activations and weights for $n = 8$, see Figure
\ref{fig:activations_and_weights}.  When testing the response for activations we
set $p = 4$ for the weights, and when testing the response for weights we set
$p = 4$ for the activations.  From Figure~\ref{fig:activations_and_weights} we
see that activations and weights both need at least a few significand bits, but
beyond that there is no significant difference in accuracies between activations
and weights or more significand bits.

\subsection{Exponent and Significand Bits}
\begin{figure}[H]
\centering
  \begin{subfigure}[b]{0.4\textwidth}
  \centering
  \includegraphics[scale=0.4]{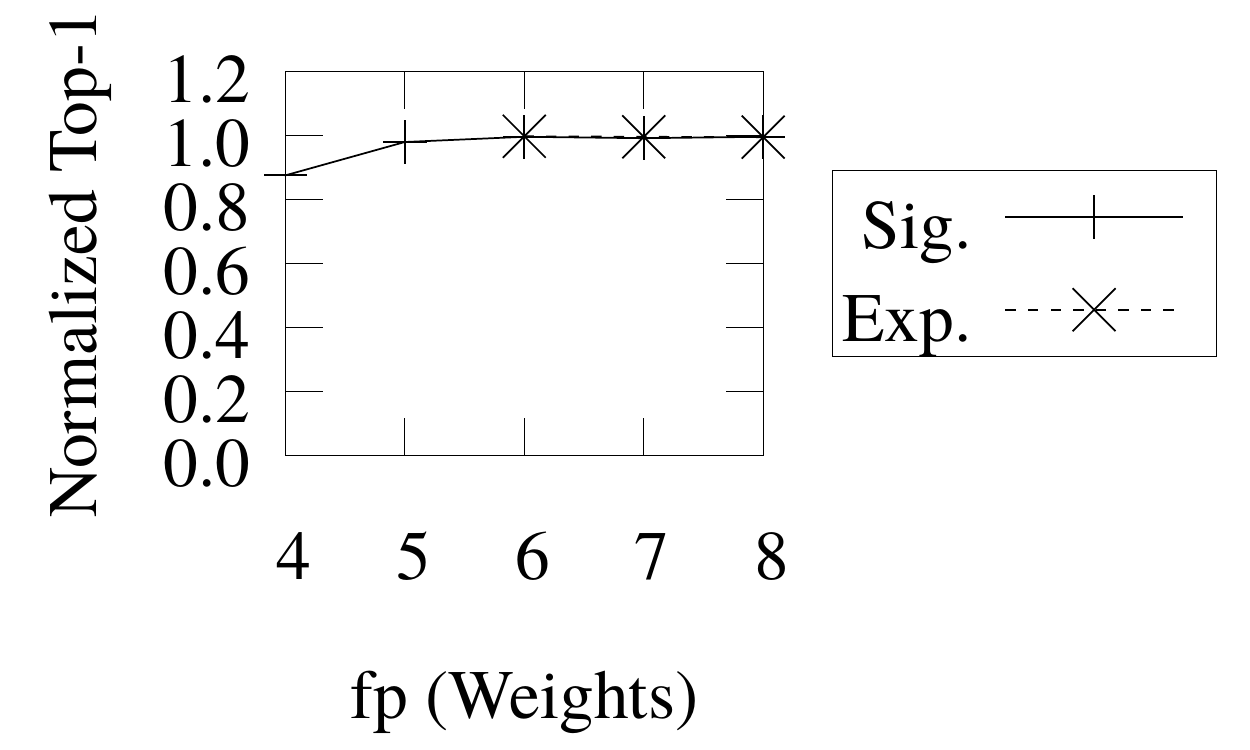}
  \caption{GoogLeNet}
  \label{fig:exponent_and_significand_bits-googlenet}
  \end{subfigure}%

  \begin{subfigure}[b]{0.4\textwidth}
  \centering
  \includegraphics[scale=0.4]{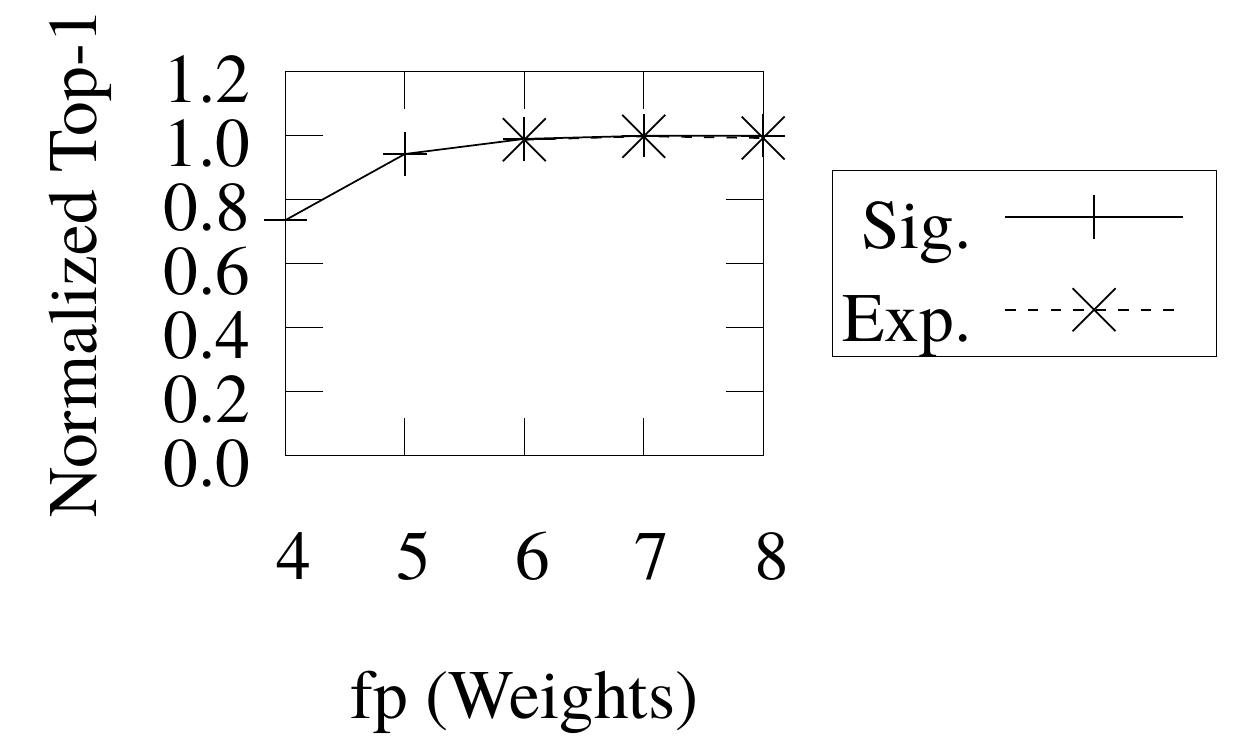}
  \caption{ResNet}
  \label{fig:exponent_and_significand_bits-resnet}
  \end{subfigure}%
~
  \begin{subfigure}[b]{0.4\textwidth}
  \centering
  \includegraphics[scale=0.4]{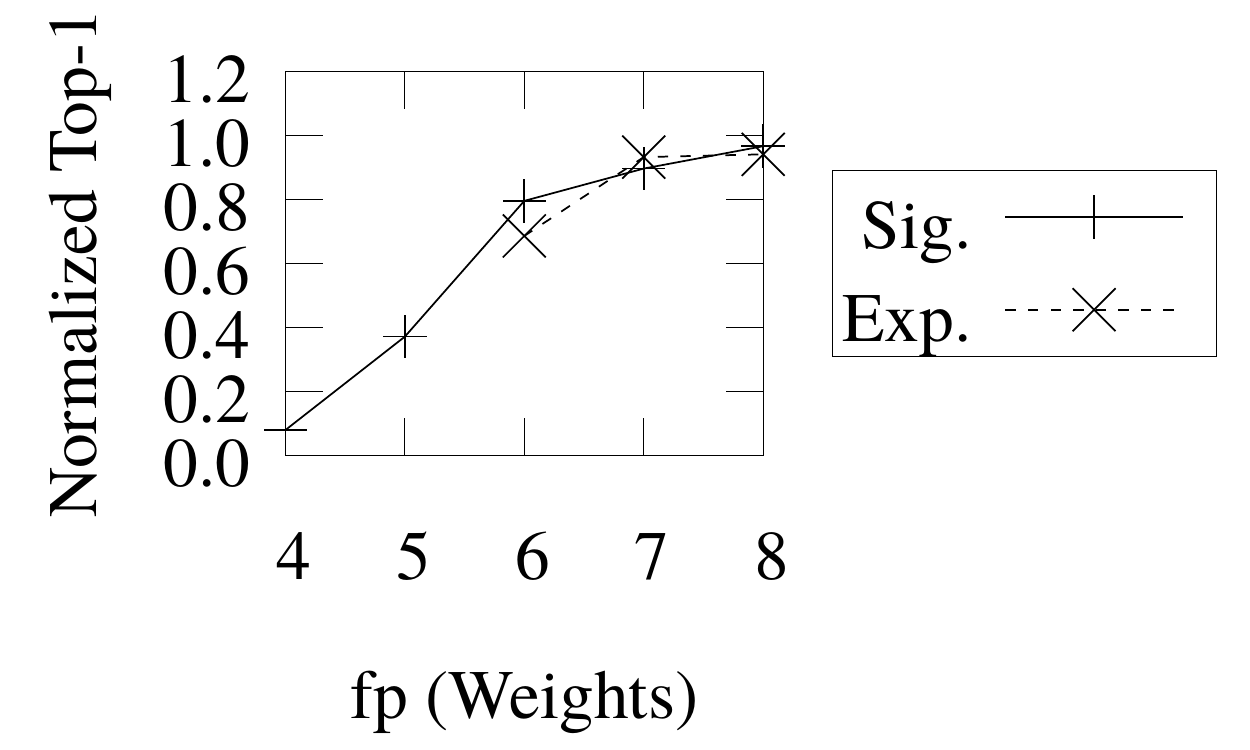}
  \caption{MobileNet}
  \label{fig:exponent_and_significand_bits-mobilenet}
  \end{subfigure}
\caption{Exponent and significand bits based on $n = 8$, $p = 4$, calibrated against 8 training images}
\label{fig:exponent_and_significand_bits}
\end{figure}

The third and final set of experiments measure the effect of reducing the
bitwidth of the weights at the cost of either exponent bits or significand
bits all while setting $n = 8$ and $p = 4$ for activations, see Figure
\ref{fig:exponent_and_significand_bits}.  When testing the response on weights
for exponent bits we start with $n = 8$ and $p = 4$ and decrement $n$ by one
until $n = 6$ all while keeping $p = 4$.  Similarly, when testing the response
on weights for significands bits we start with $n = 8$ and $p = 4$ and
decrement both $n$ and $p$ by one until $p = 0$.  In
Figure~\ref{fig:exponent_and_significand_bits} it appears that weights for
GoogLeNet and ResNet can scale down from 8 bits to only 5 bits without
significant loss in accuracies, but for MobileNet weights can only scale down
to 7 bits before loss in accuracies become prohibitive.  Furthermore, weights
appear less sensitive to the exact allocation of exponent bits versus
significand bits, but instead respond to the bitwidth itself.

\section{Discussion}\label{sec:discussion}
Our dynamic floating-point quantization scheme and procedure yields end-to-end
post quantization inference accuracies similar if not indistinguishable to
reference single-precision floating-point models by calibrating against a single inference batch, all
while shrinking hardware and energy costs by 4x or more by replacing 32-bit operations with reduced-precision operations of 8-bits or fewer.  In particular,
we achieve 100\% normalized top-1 accuracies for GoogLeNet for fp8 through fp6,
whereas Intel recently reported slightly exceeding 60\% for fp8
\cite{chui2018flexibility}.  Furthermore, we found that for fp8 the best
accuracies were obtained using $p = 4$ or $p = 3$.  Our method allows Xilinx
FPGA users to create custom circuit designs for various $n, p$ quantization
parameters and reconfigure their FPGAs to match each target network's optimal parameters.
In fact, our results show that one can achieve accurate inference results with
as few as 3 exponent bits and that 4 exponent bits is generally no better than
3 exponent bits, which would seem to
contradict a prevailing belief popularized by NVIDIA that "fp16 is great,
what's actually even better than fp16 is truncated fp32"
\cite{dally2018scaling}.  Nevertheless, cost savings are even available for a
variety of existing hardware by increasing the effective bandwidth, reducing
the effective on- and off-chip storage, even reducing the effective toggle rate
by using $p = 6$, a.k.a. int8, or $p = 2$, a.k.a. truncated half-precision
floating-point that can be casted on-the-fly to and from full half-precision
floating-point to perform arithmetic operations, albeit both at the potential
cost of inference accuracies.  We also demonstrated that for some networks fp7
and fp6 are viable alternatives to fp8 with $p = 3$ and achieved inference
accuracies superior to their dynamic fixed-point scheme analogues, which can
only be explained by the greater range and increased precision about zero
provided by our dynamic floating-point scheme.  Thus for the densest compute
platform developers may reconfigure a Xilinx FPGA for a particular $n, p$
configuration that meets their inference accuracy requirements.  Since we're
not (re)training here, our dynamic floating-point quantization scheme and
procedure can be thought of as an offline network compression with as few as
eight calibration images, which could even potentially be substituted with
artificial images obtained by playing the network in reverse.  Finally, we're
confident that we could scale down even further if allowed to re(train) with
quantization-in-the-loop (QIL) like \cite{gysel2018ristretto} given that our
accuracies are just as good or better using only a small calibration set.

\newpage

\bibliography{quantize}{}

\end{document}